\pdfoutput=1

\documentclass[11pt]{article}

\usepackage[final]{acl}

\usepackage{times}
\usepackage{latexsym}

\usepackage[T1]{fontenc}

\usepackage[utf8]{inputenc}

\usepackage{microtype}

\usepackage{inconsolata}

\usepackage{graphicx}

\usepackage{booktabs}
\usepackage{amsmath}
\usepackage{amssymb}  %
\usepackage{bm}       %
\usepackage{cleveref}
\Crefname{section}{\S}{\S\S}

\title{On Representational Dissociation of Language and Arithmetic \\in Large Language Models}

\author{
 \textbf{Riku Kisako\textsuperscript{1}},
  \textbf{Tatsuki Kuribayashi\textsuperscript{2},
 \textbf{Ryohei Sasano\textsuperscript{1}}
}
\\
 \textsuperscript{1}Nagoya University, Japan,\\
 \textsuperscript{2}Mohamed bin Zayed University of Artificial Intelligence
\\
 \{\texttt{kisako.riku.n3@s.mail, sasano@i\}.nagoya-u.ac.jp}\\
 \texttt{tatsuki.kuribayashi@mbzuai.ac.ae}
}

\begin{document}
\maketitle
\begin{abstract}
The association between language and (non-linguistic) \textit{thinking} ability in humans has long been debated, and recently, neuroscientific evidence of brain activity patterns has been considered.
Such a scientific context naturally raises an interdisciplinary question --- what about such a \textit{language-thought dissociation} in large language models (LLMs)?
In this paper, as an initial foray, we explore this question by focusing on simple arithmetic skills (e.g., \texttt{1+2=?}) as a \textit{thinking} ability and analyzing the geometry of their encoding in LLMs' representation space.
Our experiments with linear classifiers and cluster separability tests demonstrate that simple arithmetic equations and general language input are encoded in completely separated regions in LLMs' internal representation space across \textit{all} the layers, which is also supported with more controlled stimuli (e.g., spelled-out equations).
These tentatively suggest that arithmetic reasoning is mapped into a distinct region from general language input, which is in line with the neuroscientific observations of human brain activations, while we also point out their somewhat cognitively implausible geometric properties.
\end{abstract}

\section{Introduction}
It has long been questioned whether language ability in humans is associated with (non-linguistic) \textit{thinking} abilities, such as logical or social reasoning, which is traced back to philosophical debates~\cite{Davidson1975-ia,Geschwind1970-sm,Carruthers2002-fy}.
The neuroscience field has recently provided some clues to answer this question of  \textit{language-thought dissociation}~\cite{Fodorenko2013,Blank2014-tj,fedorenko2024language}; specifically, brain activities during processing linguistic and non-linguistic reasoning stimuli is different~\cite{Fedorenko2010,Hu2023-ca,Fedorenko2024}, tentatively suggesting that those abilities are dissociated at least through the lens of brain imaging.  

\begin{figure}[t]
\centering
\includegraphics[width=\linewidth]{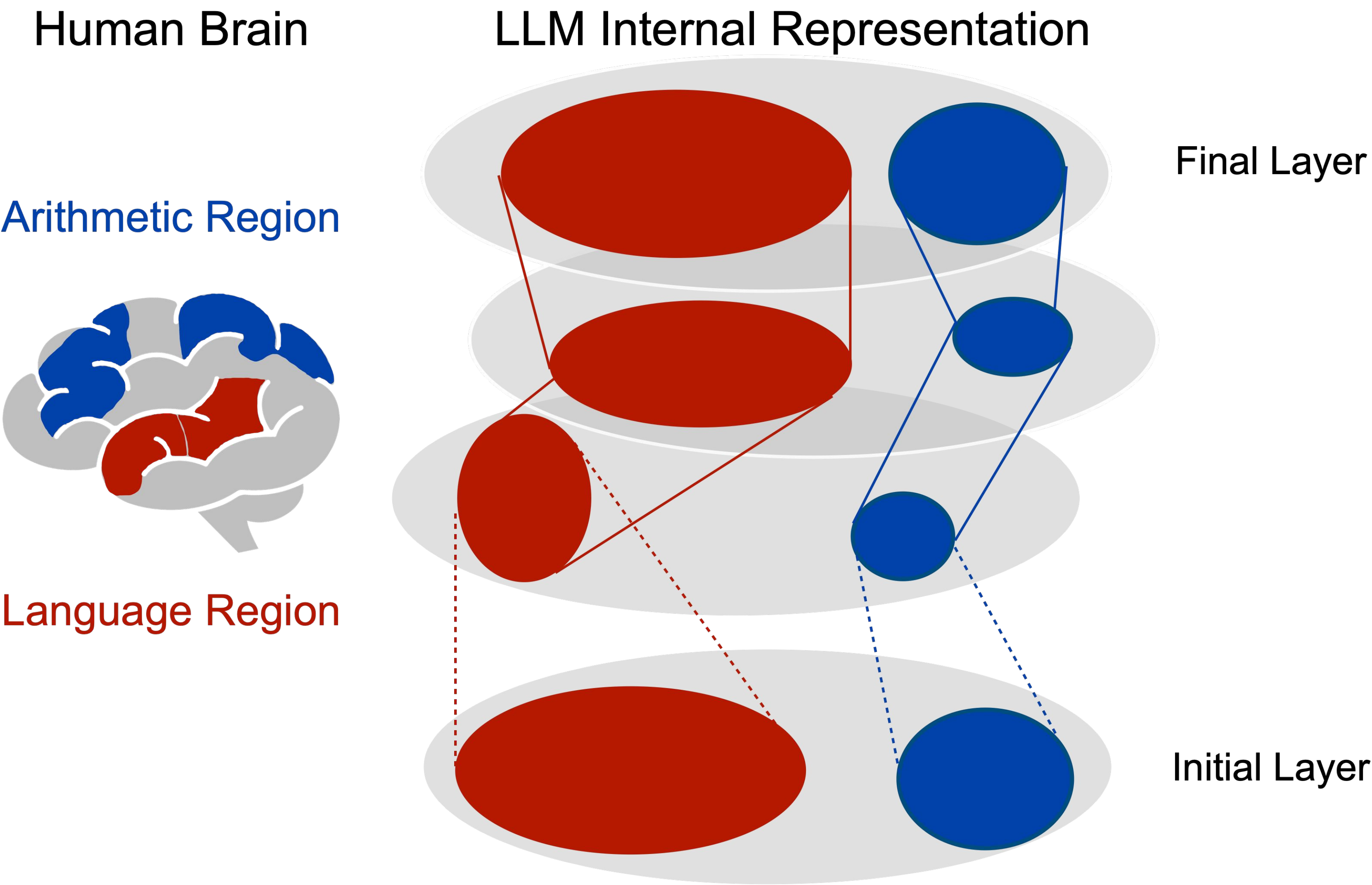}
\caption{An illustration of our perspective to investigate the language-arithmetic representational dissociation within language models (LMs) --- if brain imaging renders that the human brain activation patterns are different against linguistic and (non-linguistic) reasoning stimuli, what about LMs?}
\label{fig:fig1}
\end{figure}

Inspired by such scientific contexts, one would naturally raise an interdisciplinary question --- What about such a dissociation in the case of large language models (LLMs)? How are these (not) dissociated in LLM internals? 
There would be numerous interpretations and aspects to study regarding such a dissociation in LLMs (e.g., can one decompose LLMs into language and non-linguistic reasoning networks?), as an initial foray, 
we first quantify such a dissociation through the lens of the geometrical separation of language and arithmetic regions in the LLM's internal representation spaces, with an analogy to observing the activation patterns of human brains through brain imaging (Figure~\ref{fig:fig1}).
Notably, how LLMs encode numerical properties/operations in LLMs has been investigated~\cite{gurnee2024language,heinzerling-inui-2024-monotonic,zhu-etal-2025-language}, but the size of the associated regions and their overlap with other domains/regions have been less explored.

In our experiments, we specifically focus on arithmetic skills (e.g., \texttt{1+2=?}) as a non-linguistic \textit{thinking} ability and investigate whether these inputs are encoded in separate regions from ordinal language inputs involving general language processing.  
To address several potential confounding factors between language and arithmetic stimuli, we examined diverse and controlled data, such as spelled-out equations or texts with numerical expressions but no arithmetic operations (e.g., \texttt{Atomic number of hydrogen is ?}), clarifying LLMs' encoding strategies.
We measure the separability of language and arithmetic regions in the internal representation spaces using a linear probing classifier (\cref{subsec:setting:exp1}) and cluster separability tests (\cref{subsec:setting:exp2}).
Both experiments demonstrate that the two types of inputs are clearly separated into different clusters immediately at the first layer and never overlap throughout \textit{all} of the following layers (\cref{sec:exp}).
That is, the language and arithmetic regions are perfectly separated, at least through the lens of cluster separability in the representation space.

Nevertheless, we also observed somewhat unintuitive properties regarding language-arithmetic (excessive) representational dissociation (\cref{sec:discussion}).
Specifically, tasks requiring both language and arithmetic abilities, such as math word problems, e.g., GSM8K~\cite{cobbe2021trainingverifierssolvemath}, also formed a separate cluster from the language/arithmetic regions identified with simpler, primitive stimuli (e.g., equation) throughout the layers, instead of going back and forth between language and arithmetic regions in a pipeline-like fashion.
Furthermore, the language--arithmetic classifier trained in our main experiments (\cref{subsec:exp:linear}) always predicts the data points from math word problems to fall into language regions at \textit{all} the layers.
This suggests that the models do not treat these more complex problems as a composite of language and arithmetic problems, and the arithmetic region activated with primitive stimuli is not universally used in arithmetic-involved tasks --- arithmetic in a particular task format A (e.g., \texttt{1+2=3}) is addressed in a different region with that in a task format B (e.g., \texttt{John has one book, and he gets additional two books. How many books does John have?}).
These findings open a field to question the language-thought dissociation within LLMs.

\section{Problem setting}
\label{sec:setting}
We quantify whether/how language and arithmetic inputs are separately distributed in LLMs' internal representation space.
We adopt two approaches: linear classifier and cluster separability test.

\paragraph{Notaion}
Suppose there are two text sets: $X^{c_1}=[x_1^{c_1}, x_2^{c_1}, \cdots, x_n^{c_1}]$, and $X^{c_2}=[x_1^{c_2}, x_2^{c_2}, \cdots, x_m^{c_2}]$, where  each input $x_k$ consists of multiple tokens.
The classes $c_1$ and $c_2$ should be, for example, general language stimuli and arithmetic texts, respectively.
Our interest is how their corresponding representations $\bm H^{c_1}=[\bm h_1^{c_1}, \bm h_2^{c_1}, \cdots, \bm h_n^{c_1}] \in \mathbb{R}^{d\times n}$ and $\bm H^{c_2}=[\bm h_1^{c_2}, \bm h_2^{c_2}, \cdots, \bm h_n^{c_2}] \in \mathbb{R}^{d\times m}$ are distributed in an internal representation space $\mathbb{R}^d$ in a particular layer of an LLM.
At least in this study, we used the last token's representation $\bm h_k$  as the internal representation of $x_k$.\footnote{Most of our stimuli end with a token ``?'' to prevent confounding of the token type being encoded between language and arithmetic stimuli.}

\subsection{Liner classification}
\label{subsec:setting:exp1}
We train a linear binary classification model to distinguish representations into their associated class, motivated by the probing studies~\cite{alain2017understanding,tenney2018what,hewitt-manning-2019-structural}.
Specifically, this classifier takes a representation $\bm h  \in \bm H^{c_1} \oplus \bm H^{c_2}$ as input and predicts whether this input is from $\bm H^{c_1}$ or $\bm H^{c_2}$ (binary label),  where $\oplus$ denotes the concatenation of lists.
We used a linear SVM classifier, where the boundary between $c_1$ and $c_2$ is set based on the large-margin objective.
We split the dataset into train and test sets with 4:1.
We performed a 5-fold cross-validation and reported the average accuracy.
If the classifier achieves perfect accuracy in a held-out test set, a linear boundary hyperplane can be drawn (thus separable) between the two clusters.

\subsection{Cluster separability test}
\label{subsec:setting:exp2}
Linear classification accuracy does not tell about how \textit{distant} these two clusters are. As complementary information, we also quantify the distance between the two clusters.
We specifically adopt a generalized discrimination value (GDV)~\cite{SchillingMGMK21,kissane2025probing}, a scaling and transformation invariant measure of cluster separability.
Simply put, this measures the difference between intra-cluster and inter-cluster normalized distances: $D(c_1, c_2) = d(\bm H^{c_1}) + d(\bm H^{c_2})-d(\bm H^{c_1}, \bm H^{c_2})$.
The score $D$ of zero means that the two clusters completely overlap, and the smaller the (negative) value $D$ is, the more the clusters are distant.
GDV of -1 will already be a strong separation (see \citet{kissane2025probing} for the concise introduction of GDV).

\begin{table}[t]
\centering
\scriptsize
\tabcolsep=0.1cm
\begin{tabular}{llllp{3cm}}
\toprule
Data &  Lang. & Num. & Arith.  & Example \\
\cmidrule(r){1-5}
\textsc{Lang} & \checkmark &  & & How do you view the nature of the world we live in?\\
\cmidrule(r){1-1} \cmidrule(lr){2-4} \cmidrule(lr){5-5}
\textsc{LangNum} & \checkmark & \checkmark & & What is the atomic number of hydrogen? \\
\cmidrule(r){1-5}
\textsc{Eq} & & \checkmark & \checkmark & 3*1-2=? \\
\cmidrule(r){1-1} \cmidrule(lr){2-4} \cmidrule(lr){5-5}
\textsc{EqSp} & (\checkmark) & \checkmark & \checkmark  & three times one minus two equals? \\
\cmidrule(r){1-5}
\textsc{LangNumEq} & \checkmark & \checkmark & \checkmark & \{the number of fingers displayed in a peace sign\}-1=? \\
\cmidrule(r){1-1} \cmidrule(lr){2-4} \cmidrule(lr){5-5}
\textsc{GSM8K} & \checkmark & \checkmark & \checkmark  &
A robe takes 2 bolts of blue fiber and half that much white fiber. How many bolts in total does it take? \\
\bottomrule
\end{tabular}
\caption{Our data with an example stimuli. We list what kind of ability will be needed in each data: ``Lang.,''  ``Num.,'' and ``Arith.'' denote natural language, numerical output, and arithmetic operations, respectively.}
\label{tab:exp_data}
\end{table}

\subsection{Data}
\label{subsec:dataset}

We examine six types of data (Table~\ref{tab:exp_data}).
We explore how these stimuli are represented within LLMs.

\paragraph{General language stimuli (\textsc{Lang})}
We use 500 texts in five languages (100 texts for each language of English, Chinese, Japanese, Russian, and Arabic), sampled from a clean multilingual corpus MADLAD~\cite{kudugunta2023madlad400multilingualdocumentlevellarge}.\footnote{%
Language ability typically refers to the competence to construct meaning from linguistic structure. Through this view, a linguistic region in the human brain is often identified as activation differences between grammatical and ungrammatical (nonsensical) stimuli~\cite{Fodorenko2013}. In our cluster separability analysis, it was not obvious how to address such activation differences. In Appendix, we tentatively show that (i) grammatical and ungrammatical (word-order-shuffled) stimuli are distributed in close regions of each other and (ii) in either using grammatical or ungrammatical stimuli as \textsc{Lang}, our conclusion does not alter, i,e., these are separated from arithmetic stimuli.}
We define the language region as broadly as possible using multilingual texts, making it somewhat unlikely for the arithmetic region to be easily separated.

\paragraph{Texts about number (\textsc{LangNum})}
We also use 200 questions (100 for English and Chinese) with numerical answers (e.g., \texttt{What is the atomic number of hydrogen?}) as language stimuli without any calculation but leading to numerical output.
Including this data in the linguistic stimuli prevents us from just exploring coarse domain differences (general words vs. numbers).

\paragraph{Arithmetic equations (\textsc{Eq})}
We use 100 simple arithmetic equations (e.g., \texttt{2*2-3=?}) involving three operations (+, -, $\times$) to identify arithmetic regions.
The answer is between 1 to 10. %

\paragraph{Spelled-out equations (\textsc{EqSp})}
We spelled out the \textsc{Eq} data in English and Chinese (e.g., \texttt{two times two minus three equals?}), resulting in 200 stimuli.
This spelling-out, more or less, addresses the confounding factor of character type differences between the input of language and arithmetic stimuli (alphabet vs. number).

\vskip\baselineskip
We regard \textsc{Lang} and \textsc{LangNum} as linguistic stimuli and \textsc{Eq} and \textsc{EqSp} as arithmetic stimuli.\footnote{From the perspective of language-thought dissociation~\cite{mahowald2024dissociating}, \textsc{LangNum} might also be in a \textit{thought} domain considering these address factual knowledge. Although our research is motivated by such a broad view, at least in this study, our focus is more specific on whether arithmetic operations are included or not in the stimuli.}
We also examine more complex stimuli involving language and arithmetic processing (\cref{sec:discussion}):

\paragraph{Texts on number and arithmetic (\textsc{LangNumEq})}
We modified the 200 \textsc{LangNum} data to involve arithmetic operation (e.g., \texttt{\{the number of fingers displayed in a peace sign\}-1=?}).\footnote{The accuracy for this task is about 55\%; although it is not perfect, the model seemingly understood the problem.}

\paragraph{\textsc{GSM8K}}
We examine \textsc{GSM8K}~\cite{cobbe2021trainingverifierssolvemath}, a widely used benchmark of human-crafted math word problems at about elementary-school level.  
Solving this task requires both language processing (to first comprehend the problem written in natural language) and arithmetic operations.

\subsection{Language models}
\label{subsec:model}
We use multiple LLMs to see the generality of our results: 
Gemma-2-9b-it~\cite{gemmateam2024gemma2improvingopen}, 
Llama-3.1-8B-Instruction~\cite{grattafiori2024llama}, 
and Qwen2.5-7B-Instruct~\cite{qwen2024qwen25}.
The results for Gemma-2-9b-it are demonstrated in the following sections (see Appendix for other models' results).

\section{Experiments}
\label{sec:exp}

\subsection{Linear separability}
\label{subsec:exp:linear}
\paragraph{Setting}
We first examine the classification between \textsc{Lang}$\oplus$\textsc{LangNum} (language) and \textsc{Eq}$\oplus$\textsc{EqSp} (arithmetic) as a general language and arithmetic dissociation.

\paragraph{Results}
The classification accuracy achieved 100\%  at all the layers except for the embedding layer.
This suggests that language and arithmetic regions are immediately separated from the first layer (representationally dissociated), and the exception in the embedding layer rules out the possibility of this separation being merely from the token difference.
Our results entail the separation of \textsc{LangNum} and \textsc{Eq}, indicating that there is a specific region to arithmetic operations, not just for the numerical domain.
This also entails the separation of \textsc{Lang}/\textsc{LangNum} and \textsc{EqSp}; the \textsc{EqSp} data is spelled out, and that dissociation cannot be explained merely by the token type difference (words vs. numerical characters).
Notably, the intra-language/arithmetic classifications of \textsc{Lang} vs. \textsc{LangNum} and \textsc{Eq} vs. \textsc{EqSp} were also be possible with 100\% accuracies, showing that each set of stimuli forms smaller clusters even within the arithmetic and language regions.
Appendix~\ref{sec:appendix:result_pca} also provides PCA visualizations of their separations.

\subsection{Cluster distances}
\label{subsec:exp:cluster}
\paragraph{Setting}
To further interpret the geometrical distance between the clusters, we computed the GDV scores for particular pairs of datasets.

\paragraph{Results}
The upper part of Figure~\ref{fig:exp2} shows the GDV scores between language--arithmetic cluster pairs for Gemma-2 (see \autoref{sec_appendix:cluster_result_other_models} for other LLMs' results).
First of all, the scores are negative in most pairs and layers, supporting our previous findings of the representational dissociation between language and arithmetic.
There are no distinctive trends in the similarity of stimuli, and all the pairs are equally distant from each other, e.g., \textsc{Land}--\textsc{Eq} and \textsc{Land}--\textsc{EqSp} are equally distant independent of superficial character similarities (character vs. numbers).

\section{Analysis: language-arithmetic combined tasks}
\label{sec:discussion}
\begin{figure}[t]
\centering
\includegraphics[width=0.90\linewidth]{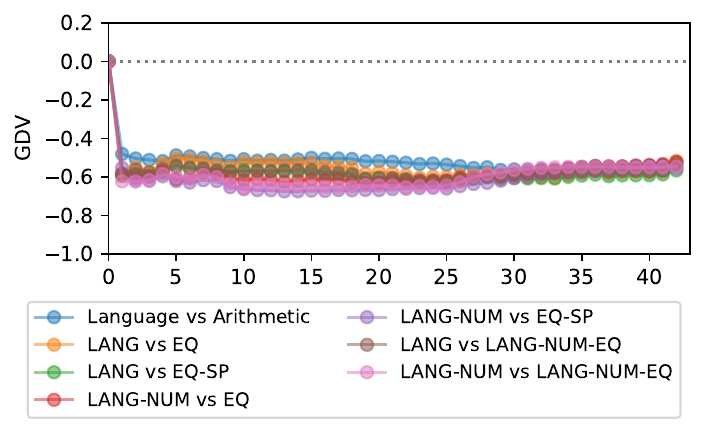}
\includegraphics[width=0.90\linewidth]{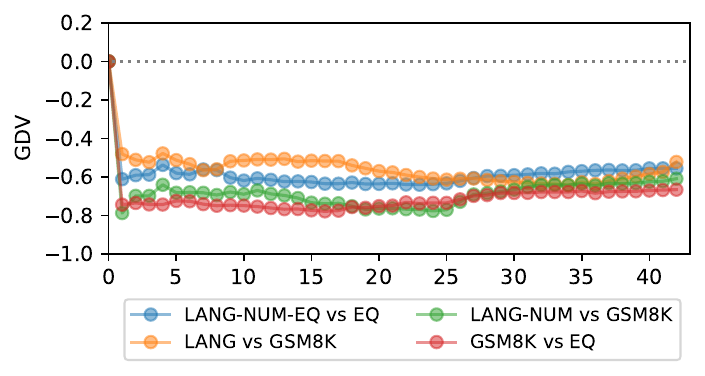}
\caption{The GDV between clusters of interest (for Gemma-2). ``Language vs.\ Arithmetic'' is the distance between \textsc{Lang}$\oplus$\textsc{LangNum} and \textsc{Eq}$\oplus$\textsc{EqSp}.}
\label{fig:exp2}
\end{figure}

We identified language and arithmetic regions with primitive stimuli (\textsc{Lang}, \textsc{LangNum}, \textsc{Eq}, \textsc{EqSp}) in~\cref{{sec:exp}}.
Given that LLMs are typically employed to solve more complex tasks, such as reading comprehension problems, we investigate how LLMs encode language-arithmetic combined problems, such as math word problems (\textsc{LangNumEq} and \textsc{GSM8K}).
Intuitively, such tasks require a pipeline processing of (i) first comprehending the text (e.g., ``Tom gets two books''  to be \texttt{Tom++2}), and then (ii) performing arithmetic operations to derive an answer (e.g., \texttt{Tom=0, exec(Tom++2), Tom=2}).
This expects, more or less, the corresponding representations to go back and forth between language and arithmetic regions.
In contrast, we observed: (i) \textsc{LangNumEq} and \textsc{GSM8K} inputs always fall into language regions based on the linear classifier trained in~\cref{subsec:exp:linear}, and (ii) these data points also form a distinct cluster, which never overlap with language nor arithmetic through the lens of cluster separability~\cref{subsec:exp:cluster} (see the bottom part of Figure~\ref{fig:exp2}).
This suggests that the models do not treat these more complex problems as a composite of language and arithmetic problems, at least through the lens of our representational analysis, and thus, the arithmetic region activated with primitive stimuli (e.g., \textsc{Eq}) was no longer used in particular more complex arithmetic-involved tasks (\textsc{LangNumEq} and \textsc{GSM8K}).
This suggests that there are numerous arithmetic regions for different task formats instead of a single universal arithmetic region; in other words, LLMs' representation space has too fine-grained modularity beyond a course language--arithmetic dissociation.

\section{Conclusion}
In this study, we have explored the representational dissociation between language and arithmetic stimuli, inspired by the neuroscientific observation of brain activation differences in language and non-linguistic reasoning processing.
Our work sets the direction to bridge/contextualize the neuroscientific perspective of language and thought dissociation with LLM's internal analysis.

\section*{Limitations}
Numerous findings have been accumulated on the geometry of LLM's representation space; for example, on its task-specific subspaces~\cite{aghajanyan-etal-2021-intrinsic,zhang-etal-2023-fine,weber-etal-2024-interpretability}, the existence of neurons specific to particular knowledge/domain~\cite{dai-etal-2022-knowledge,tang-etal-2024-language,kojima-etal-2024-multilingual}, outlier dimensions~\cite{kovaleva-etal-2021-bert,puccetti-etal-2022-outlier,yu2025the}, or its semantic stability~\cite{wu2024semantic}.
Incorporating such perspectives into our analysis is our future work.
In particular, there would be alternative approaches to identify the language and arithmetic regions via, e.g., gradient descent~\cite{weber-etal-2024-interpretability}, the magnitude of neuron activations~\cite{kojima-etal-2024-multilingual}, or activation probability~\cite{tang-etal-2024-language}.
The generality of our findings can be enhanced with other methodologies as well. 
One important point is that we just \textit{observed} the geometrical property of the internal representation space, and whether our observation is aligned with LLM's actual abilities/behaviors should be tested via causal analysis~\cite{amini-etal-2023-naturalistic,stolfo-etal-2023-mechanistic}.
For example, if one claims that language and arithmetic abilities are dissociated within LLMs, one should be able to design interventions that only hurt linguistic knowledge (fluency) or specific reasoning abilities of LLMs.
Such a demonstration will serve as stronger evidence.

Another interesting direction is to contextualize our results with the existing LM-brain alignment studies~\cite{Kumar2024-ma,aw2024instructiontuning}.
Such connection will probably be enhanced by incorporating actual human neuroimaging data into the analysis.

Lastly, there are several aspects of the language-thought dissociation argument.
We just analyzed the internals of pre-trained models, but this topic should also be relevant to the learnability of specific abilities.
For example, does achieving good reasoning skills pre-require good language skills? Such a related topic should be addressed through different experimental designs, e.g., ablating the language learning scenarios~\cite{warstadt2022what}.

\section*{Ethics Statement}
This work is solely on the analysis of LLMs and does not involve, e.g., potentially harmful/sensitive data, human-involved experiments, or the construction of new data.
We used AI assistance tools within the scope of “Assistance purely with the language of the paper” described in the ACL 2023 Policy on AI Writing Assistance.

\clearpage
\appendix
\section{SVM Hyperparameter Settings}
For the linear classification experiments, we employed scikit-learn’s implementation of linear SVM (i.e., SVC), which utilizes the libsvm library. The hyperparameters were set as follows: 
\begin{itemize}
    \item Kernel: \textit{linear}
    \item Regulrization parameter (C): $1.0$
    \item Tolerance (tol): $1e^{-3}$
    \item Maximum iterations (max\_iter): $-1$ (no limit)
\end{itemize}

\section{Model details}
The license of the used models/data is listed in \autoref{tab:licence}; all of them are used under their intended use.
All models were accessed using the Hugging Face toolkit\cite{wolf-etal-2020-transformers} and tested on a single NVIDIA A100 GPU (80GB). All experiments were conducted within 100 GPU hours.

\begin{table}[h]
    \centering
    \tabcolsep=0.1cm
    \begin{tabular}{llp{3cm}}
    \toprule
         Data/Model&Licence  \\
         \cmidrule(r){1-1}\cmidrule(r){2-2}
         MADLAD-400& Creative Commons CC-BY-4.0\\
         GSM8K&MIT\\
         Gemma2&Gemma Terms of Use License\\
         Llama3.1&Llama 3.1 Community License\\
         Qwen2.5&Apache license 2.0\\
    \bottomrule
    \end{tabular}
    \caption{Licence of the data and models}
    \label{tab:licence}
\end{table}

\section{Prompting}
\autoref{tab:prompt} shows the exact prompt format we employed for each dataset.
For \textsc{Lang}, we entered the input text as is, and for \textsc{LangNum}, \textsc{Eq},\textsc{EqSp},\textsc{LangNumEq} and \textsc{GSM8K}, we explicitly instructed the model to answer the \textit{numbers} to avoid potential confounding of output token types.
\begin{table}[t]
\centering
\scriptsize
\tabcolsep=0.1cm
\begin{tabular}{llp{3cm}}
\toprule
Data & Prompt \\
\cmidrule(r){1-2}
\begin{tabular}{l}
     \textsc{Lang}
\end{tabular}
&
\verb|<INPUT_TEXT>|
\\
\cmidrule(r){1-2}
\begin{tabular}{l}
     \textsc{LangNum}\\
     \textsc{Eq} \\
     \textsc{EqSp}\\
     \textsc{LangNumEq}\\
     \textsc{GSM8K}
\end{tabular}
& 
\begin{tabular}{l}
\verb|Please answer the following question by|\\
\verb|providing numbers alone as your answer|\\
\verb|:<INPUT_TEXT>|
\end{tabular}
\\
\bottomrule
\end{tabular}
    \caption{The exact prompts we employed for each dataset}
    \label{tab:prompt}
\end{table}

\section{Liner classification results}
\label{sec_appendix:liner_classification}
\begin{figure}[h]
    \centering
    \includegraphics[width=0.70\linewidth]{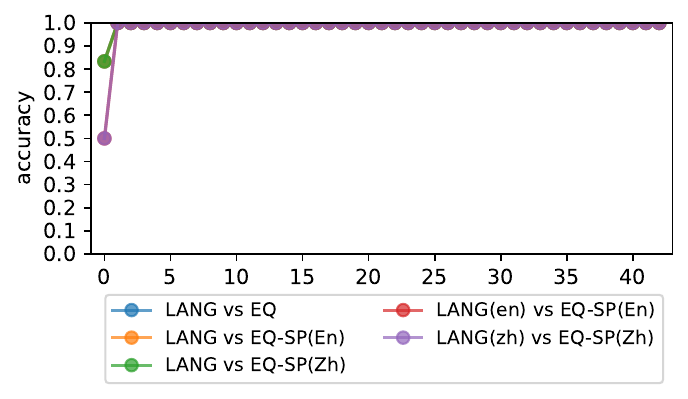}
    \includegraphics[width=0.70\linewidth]{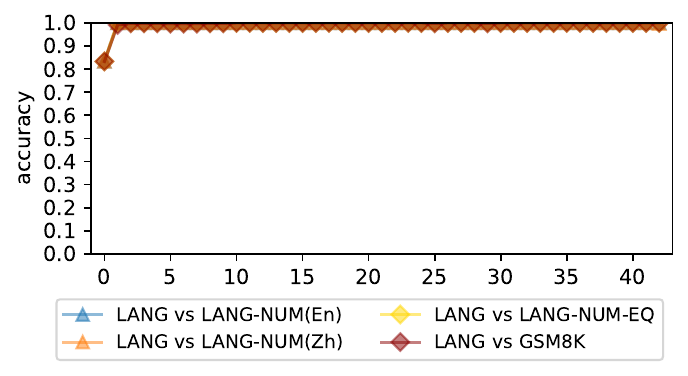}
    \includegraphics[width=0.70\linewidth]{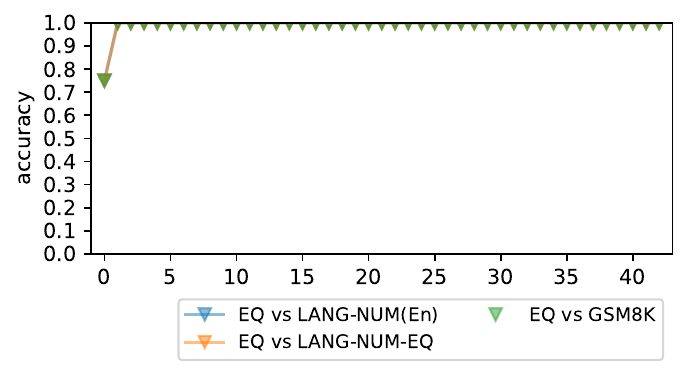}
    \caption{Liner classification results of gemma-2-9b-it}
    \label{fig:gemma-liner}
\end{figure}

\begin{figure}
    \centering
    \includegraphics[width=0.70\linewidth]{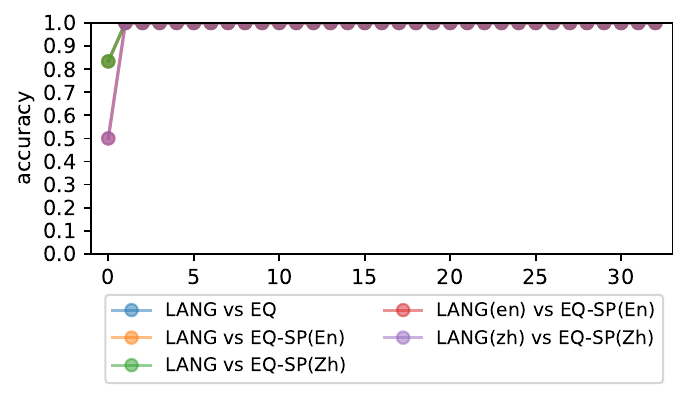}
    \includegraphics[width=0.70\linewidth]{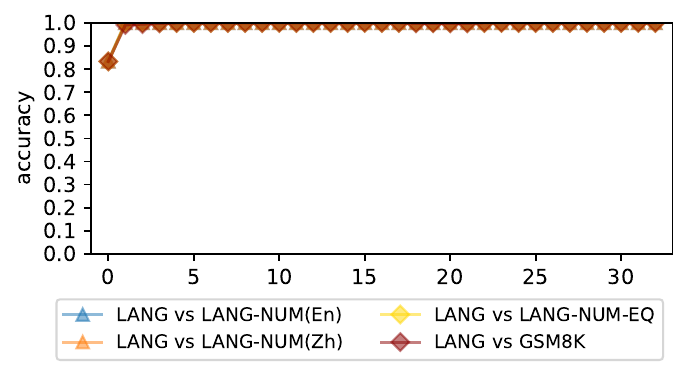}
    \includegraphics[width=0.70\linewidth]{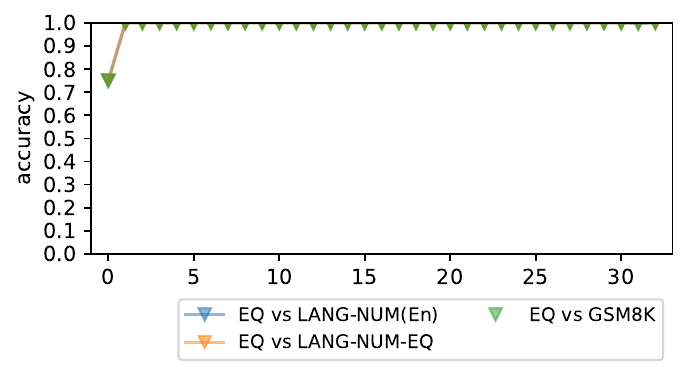}
    \caption{Liner classification results of Llama-3.1-8B-Instruction}
    \label{fig:llama-liner}
\end{figure}

\begin{figure}
    \centering
    \includegraphics[width=0.70\linewidth]{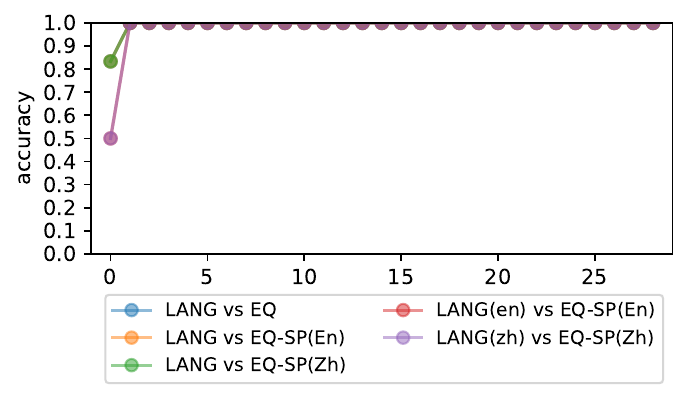}
    \includegraphics[width=0.70\linewidth]{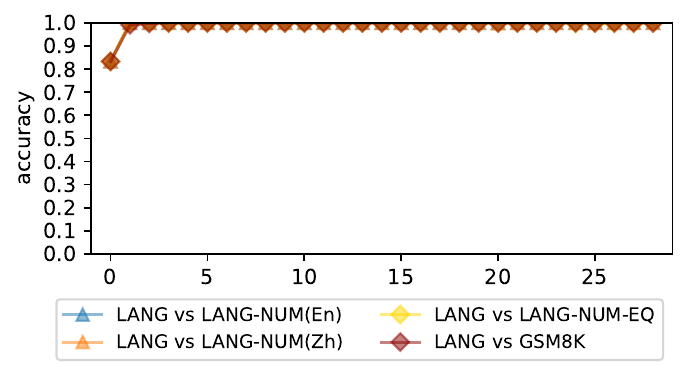}
    \includegraphics[width=0.70\linewidth]{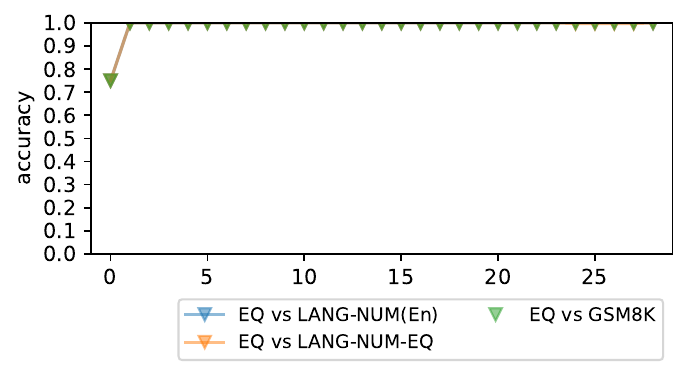}
    \caption{Liner classification results of Qwen2.5-7B-Instruct}
    \label{fig:qwen-liner}
\end{figure}

\autoref{fig:gemma-liner}, \autoref{fig:llama-liner} and \autoref{fig:qwen-liner} show the liner classification results for gemma-2-9b-it, Llama-3.1-8B-Instruction and Qwen2.5-7B-Instruct, respectively.


\section{Cluster distances results of other models}
\label{sec_appendix:cluster_result_other_models}
\begin{figure}[t]
    \centering
    \includegraphics[width=0.93\linewidth]{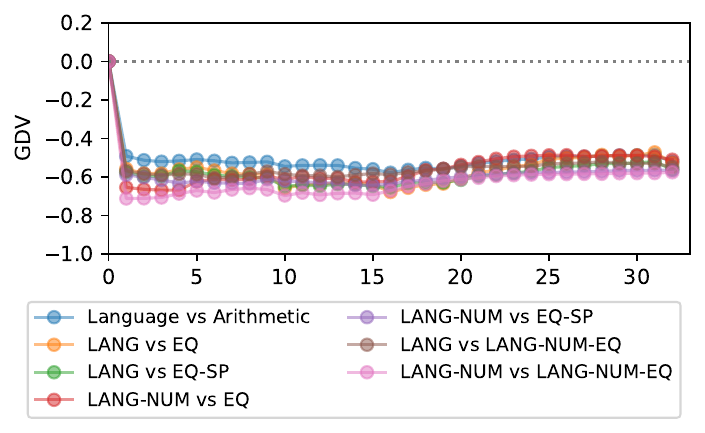}
    \includegraphics[width=0.93\linewidth]{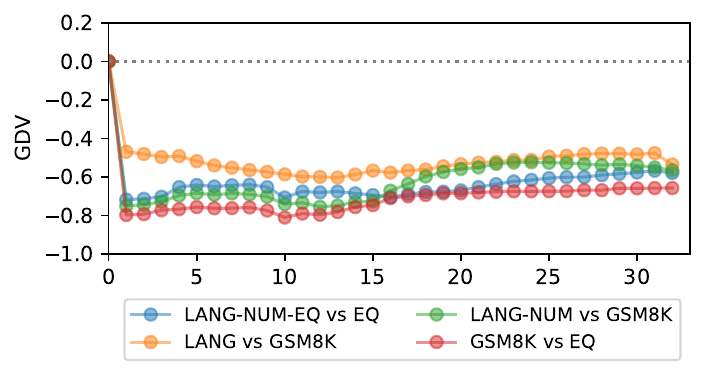}
    \caption{The GDV scores of Llama-3.1-8B-Instruction}
    \label{fig:llama_cluster}
    \hspace{0.1\columnwidth}
    \centering
    \includegraphics[width=0.93\linewidth]{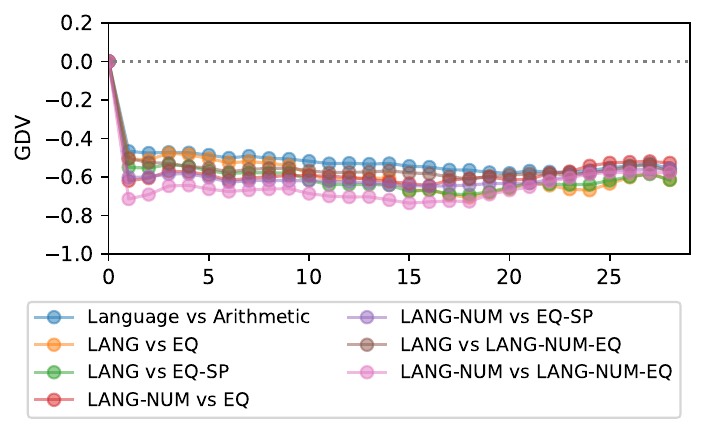}
    \includegraphics[width=0.93\linewidth]{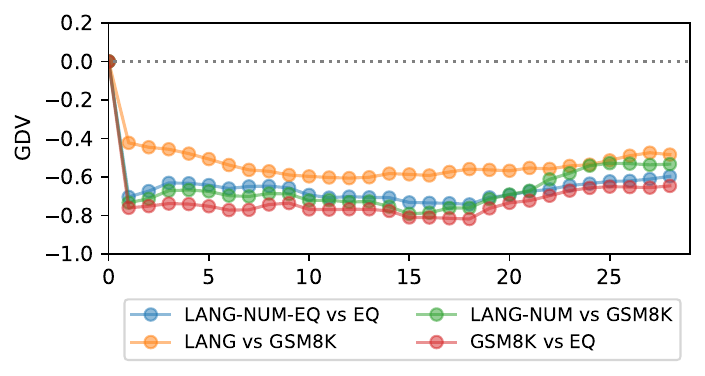}
    \caption{The GDV scores of Qwen2.5-7B-Instruct}
    \label{fig:qwen_cluster}
\end{figure}

To see the generality of the results in \cref{subsec:exp:cluster} across different LLMs, we also examined Llama-3.1-8B-Instruction and Qwen2.5-7B-Instruct.
\autoref{fig:llama_cluster} and \autoref{fig:qwen_cluster} show the GDV results (\cref{subsec:exp:cluster}) for Llama-3.1-8B-Instruction and Qwen2.5-7B-Instruct, respectively.
All the results show negative GDV scores, which is consistent with our main findings.

\section{Visualizations of internal representation space}
\label{sec:appendix:result_pca}
\begin{figure}[h]
    \centering
    \includegraphics[width=0.93\linewidth]{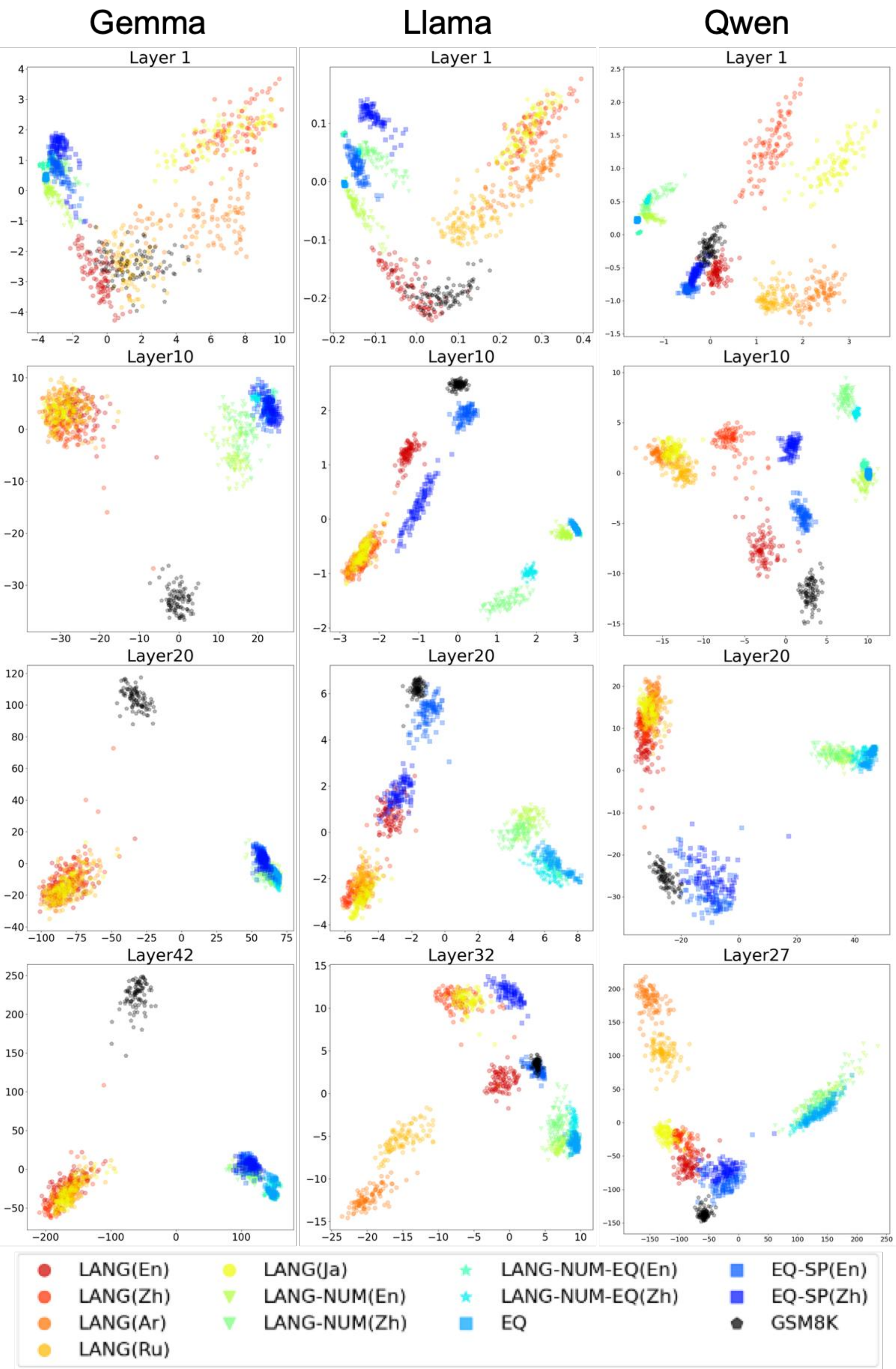}
    \caption{2D visualization of internal representation space via PCA. The results for three models are shown (Gemma, Llama, and Qwen).}
    \label{fig:pca_result}
\end{figure}

Figure~\ref{fig:pca_result} shows the 2D visualization of internal representation space using PCA. Layer 1, 10, 20, and the final layers are sampled. 

\section{Word-order-shuffled language stimuli}
\label{sec:appendix:language_data}
\begin{figure}[h]
    \centering
    \includegraphics[width=0.8\linewidth]{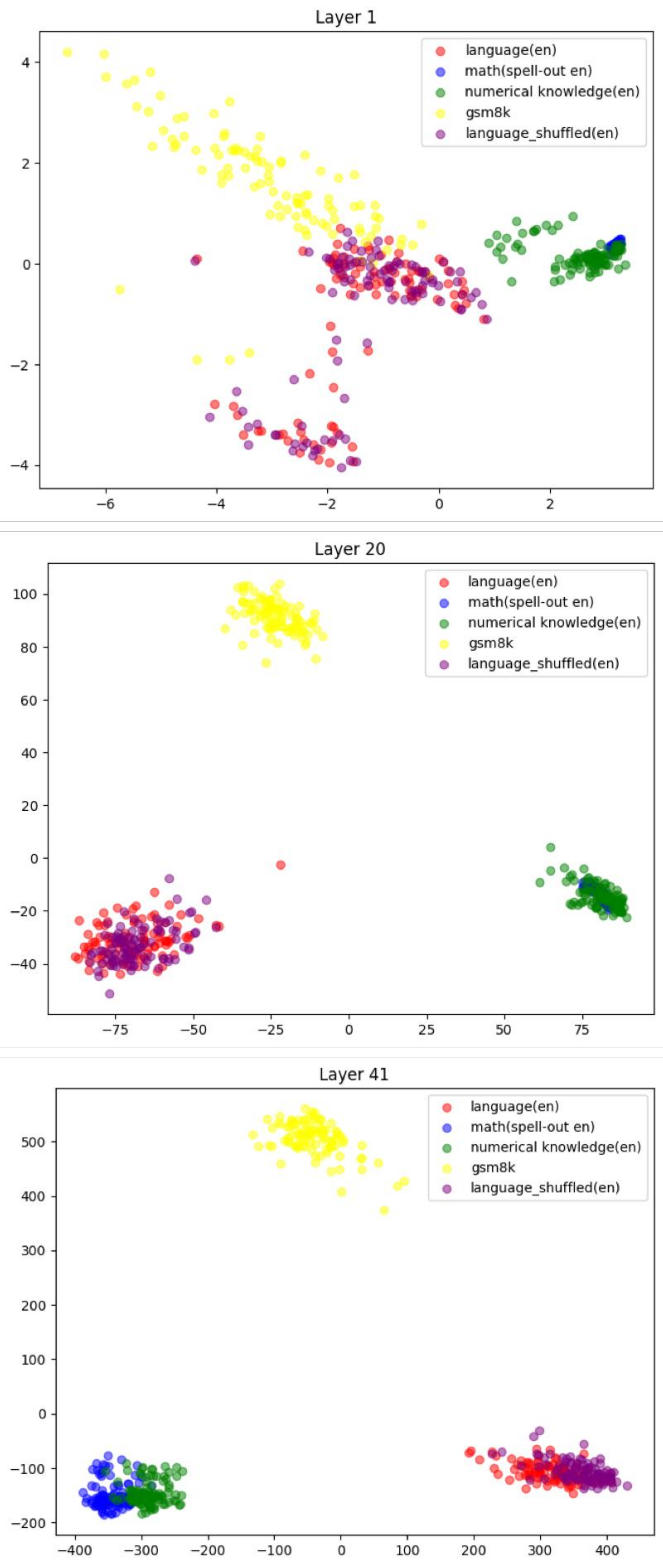}
    \caption{The 2D visualization of \textsc{Lang} (en), \textsc{LangNum} (en), \textsc{EqSp} (en), \textsc{GSM8K}, and \textsc{LangShuffled} (en) clusters in Gemma2-9b-it via PCA.}
    \label{fig:pca_result_shuffle}
\end{figure}

Given the view that language ability refers to the competence to construct meaning from linguistic structure, existing studies typically use word-order shuffled sentences as (ungrammatical) baseline stimuli, while we did not directly incorporate such a perspective into our experimental design.
Here, we also examine such stimuli \textsc{LangShuffled} in our experimental design and show that  (i) grammatical and ungrammatical (word-order-shuffled) stimuli are distributed in close regions of each other and (ii) in either using grammatical or ungrammatical stimuli, our conclusion does not alter, i,e., these are separated from arithmetic stimuli.
Figure~\ref{fig:pca_result_shuffle} showcases that, regardless of the choice of grammatical/ungrammatical stimuli (\textsc{Lang}/\textsc{LangShuffled}), these are distant from arithmetic regions.
Note that we incidentally observed that the \textsc{Lang} and \textsc{LangShuffled} clusters are gradually separated through layers; that is, the model successfully encodes the grammaticality of stimuli in different regions eventually (again, both of them are distant from arithmetic regions, though).

\begin{figure}[h]
    \centering
    \includegraphics[width=0.8\linewidth]{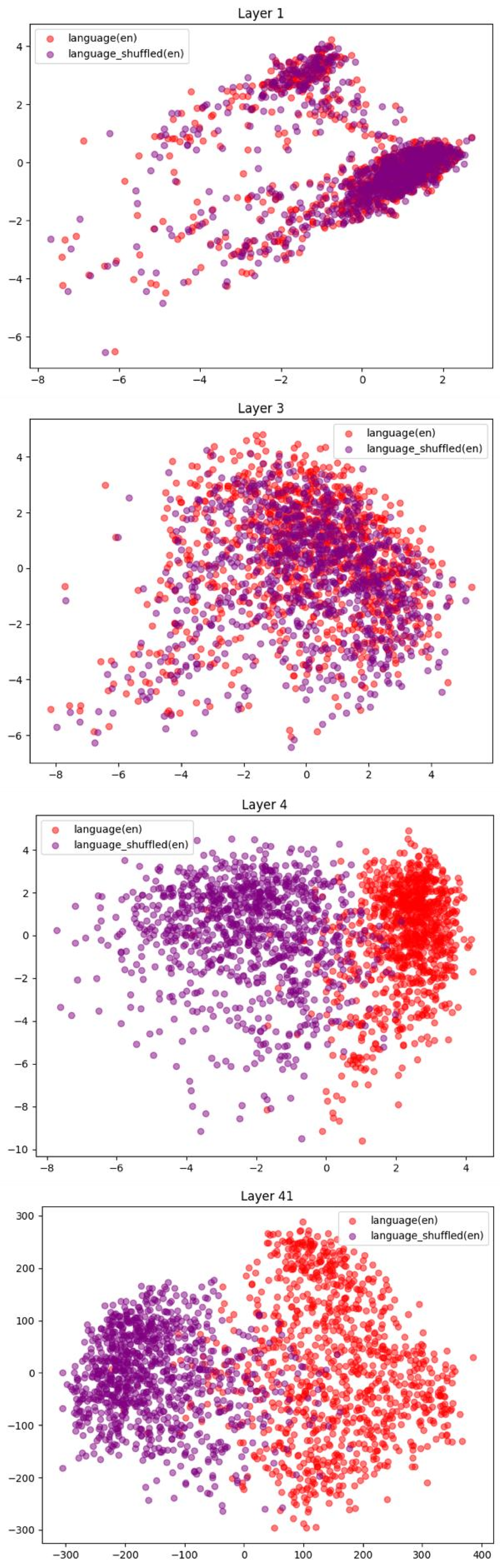}
    \caption{The 2D visualization of \textsc{Lang} (en) and \textsc{LangShuffled} (en) clusters in Gemma2-9b-it via PCA.}
    \label{fig:pca_result_only_lang}
\end{figure}


\begin{thebibliography}{39}
\providecommand{\natexlab}[1]{#1}

\bibitem[{Aghajanyan et~al.(2021)Aghajanyan, Gupta, and Zettlemoyer}]{aghajanyan-etal-2021-intrinsic}
Armen Aghajanyan, Sonal Gupta, and Luke Zettlemoyer. 2021.
\newblock \href {https://doi.org/10.18653/v1/2021.acl-long.568} {Intrinsic dimensionality explains the effectiveness of language model fine-tuning}.
\newblock In \emph{Proceedings of the 59th Annual Meeting of the Association for Computational Linguistics and the 11th International Joint Conference on Natural Language Processing (Volume 1: Long Papers)}, pages 7319--7328, Online. Association for Computational Linguistics.

\bibitem[{Alain and Bengio(2017)}]{alain2017understanding}
Guillaume Alain and Yoshua Bengio. 2017.
\newblock \href {https://openreview.net/forum?id=ryF7rTqgl} {Understanding intermediate layers using linear classifier probes}.

\bibitem[{Amini et~al.(2023)Amini, Pimentel, Meister, and Cotterell}]{amini-etal-2023-naturalistic}
Afra Amini, Tiago Pimentel, Clara Meister, and Ryan Cotterell. 2023.
\newblock \href {https://doi.org/10.1162/tacl_a_00554} {Naturalistic causal probing for morpho-syntax}.
\newblock \emph{Transactions of the Association for Computational Linguistics}, 11:384--403.

\bibitem[{Aw et~al.(2024)Aw, Montariol, AlKhamissi, Schrimpf, and Bosselut}]{aw2024instructiontuning}
Khai~Loong Aw, Syrielle Montariol, Badr AlKhamissi, Martin Schrimpf, and Antoine Bosselut. 2024.
\newblock \href {https://openreview.net/forum?id=nXNN0x4wbl} {Instruction-tuning aligns {LLM}s to the human brain}.
\newblock In \emph{First Conference on Language Modeling}.

\bibitem[{Blank et~al.(2014)Blank, Kanwisher, and Fedorenko}]{Blank2014-tj}
Idan Blank, Nancy Kanwisher, and Evelina Fedorenko. 2014.
\newblock A functional dissociation between language and multiple-demand systems revealed in patterns of {BOLD} signal fluctuations.
\newblock \emph{J. Neurophysiol.}, 112(5):1105--1118.

\bibitem[{Carruthers(2002)}]{Carruthers2002-fy}
Peter Carruthers. 2002.
\newblock The cognitive functions of language.
\newblock \emph{Behav. Brain Sci.}, 25(6):657--674.

\bibitem[{Cobbe et~al.(2021)Cobbe, Kosaraju, Bavarian, Chen, Jun, Kaiser, Plappert, Tworek, Hilton, Nakano, Hesse, and Schulman}]{cobbe2021trainingverifierssolvemath}
Karl Cobbe, Vineet Kosaraju, Mohammad Bavarian, Mark Chen, Heewoo Jun, Lukasz Kaiser, Matthias Plappert, Jerry Tworek, Jacob Hilton, Reiichiro Nakano, Christopher Hesse, and John Schulman. 2021.
\newblock \href {https://arxiv.org/abs/2110.14168} {Training verifiers to solve math word problems}.
\newblock \emph{Preprint}, arXiv:2110.14168.

\bibitem[{Dai et~al.(2022)Dai, Dong, Hao, Sui, Chang, and Wei}]{dai-etal-2022-knowledge}
Damai Dai, Li~Dong, Yaru Hao, Zhifang Sui, Baobao Chang, and Furu Wei. 2022.
\newblock \href {https://doi.org/10.18653/v1/2022.acl-long.581} {Knowledge neurons in pretrained transformers}.
\newblock In \emph{Proceedings of the 60th Annual Meeting of the Association for Computational Linguistics (Volume 1: Long Papers)}, pages 8493--8502, Dublin, Ireland. Association for Computational Linguistics.

\bibitem[{Davidson(1975)}]{Davidson1975-ia}
Donald Davidson. 1975.
\newblock Thought and talk.

\bibitem[{Fedorenko et~al.(2011)Fedorenko, Behr, and Kanwisher}]{Fodorenko2013}
Evelina Fedorenko, Michael~K. Behr, and Nancy Kanwisher. 2011.
\newblock \href {https://doi.org/10.1073/pnas.1112937108} {Functional specificity for high-level linguistic processing in the human brain}.
\newblock \emph{Proceedings of the National Academy of Sciences}, 108(39):16428--16433.

\bibitem[{Fedorenko et~al.(2010)Fedorenko, Hsieh, Nieto-Casta\~{n}\'{o}n, Whitfield-Gabrieli, and Kanwisher}]{Fedorenko2010}
Evelina Fedorenko, Po-Jang Hsieh, Alfonso Nieto-Casta\~{n}\'{o}n, Susan Whitfield-Gabrieli, and Nancy Kanwisher. 2010.
\newblock \href {https://doi.org/10.1152/jn.00032.2010} {New method for fmri investigations of language: Defining rois functionally in individual subjects}.
\newblock \emph{Journal of Neurophysiology}, 104(2):1177--1194.
\newblock PMID: 20410363.

\bibitem[{Fedorenko et~al.(2024{\natexlab{a}})Fedorenko, Ivanova, and Regev}]{fedorenko2024language}
Evelina Fedorenko, Anna~A Ivanova, and Tamar~I Regev. 2024{\natexlab{a}}.
\newblock The language network as a natural kind within the broader landscape of the human brain.
\newblock \emph{Nature Reviews Neuroscience}, pages 1--24.

\bibitem[{Fedorenko et~al.(2024{\natexlab{b}})Fedorenko, Piantadosi, and Gibson}]{Fedorenko2024}
Evelina Fedorenko, Steven~T. Piantadosi, and Edward A.~F. Gibson. 2024{\natexlab{b}}.
\newblock \href {https://doi.org/10.1038/s41586-024-07522-w} {Language is primarily a tool for communication rather than thought}.
\newblock \emph{Nature}, 630(8017):575--586.

\bibitem[{Geschwind(1970)}]{Geschwind1970-sm}
N~Geschwind. 1970.
\newblock The organization of language and the brain.
\newblock \emph{Science}, 170(3961):940--944.

\bibitem[{Gurnee and Tegmark(2024)}]{gurnee2024language}
Wes Gurnee and Max Tegmark. 2024.
\newblock \href {https://openreview.net/forum?id=jE8xbmvFin} {Language models represent space and time}.
\newblock In \emph{The Twelfth International Conference on Learning Representations}.

\bibitem[{Heinzerling and Inui(2024)}]{heinzerling-inui-2024-monotonic}
Benjamin Heinzerling and Kentaro Inui. 2024.
\newblock \href {https://doi.org/10.18653/v1/2024.acl-short.18} {Monotonic representation of numeric attributes in language models}.
\newblock In \emph{Proceedings of the 62nd Annual Meeting of the Association for Computational Linguistics (Volume 2: Short Papers)}, pages 175--195, Bangkok, Thailand. Association for Computational Linguistics.

\bibitem[{Hewitt and Manning(2019)}]{hewitt-manning-2019-structural}
John Hewitt and Christopher~D. Manning. 2019.
\newblock \href {https://doi.org/10.18653/v1/N19-1419} {{A} structural probe for finding syntax in word representations}.
\newblock In \emph{Proceedings of the 2019 Conference of the North {A}merican Chapter of the Association for Computational Linguistics: Human Language Technologies, Volume 1 (Long and Short Papers)}, pages 4129--4138, Minneapolis, Minnesota. Association for Computational Linguistics.

\bibitem[{Hu et~al.(2023)Hu, Small, Kean, Takahashi, Zekelman, Kleinman, Ryan, Nieto-Castañón, Ferreira, and Fedorenko}]{Hu2023-ca}
Jennifer Hu, Hannah Small, Hope Kean, Atsushi Takahashi, Leo Zekelman, Daniel Kleinman, Elizabeth Ryan, Alfonso Nieto-Castañón, Victor Ferreira, and Evelina Fedorenko. 2023.
\newblock Precision {fMRI} reveals that the language-selective network supports both phrase-structure building and lexical access during language production.
\newblock \emph{Cereb. Cortex}, 33(8):4384--4404.

\bibitem[{Kissane et~al.(2025)Kissane, Schilling, and Krauss}]{kissane2025probing}
Hassane Kissane, Achim Schilling, and Patrick Krauss. 2025.
\newblock Probing internal representations of multi-word verbs in large language models.
\newblock \emph{arXiv preprint arXiv:2502.04789}.

\bibitem[{Kojima et~al.(2024)Kojima, Okimura, Iwasawa, Yanaka, and Matsuo}]{kojima-etal-2024-multilingual}
Takeshi Kojima, Itsuki Okimura, Yusuke Iwasawa, Hitomi Yanaka, and Yutaka Matsuo. 2024.
\newblock \href {https://doi.org/10.18653/v1/2024.naacl-long.384} {On the multilingual ability of decoder-based pre-trained language models: Finding and controlling language-specific neurons}.
\newblock In \emph{Proceedings of the 2024 Conference of the North American Chapter of the Association for Computational Linguistics: Human Language Technologies (Volume 1: Long Papers)}, pages 6919--6971, Mexico City, Mexico. Association for Computational Linguistics.

\bibitem[{Kovaleva et~al.(2021)Kovaleva, Kulshreshtha, Rogers, and Rumshisky}]{kovaleva-etal-2021-bert}
Olga Kovaleva, Saurabh Kulshreshtha, Anna Rogers, and Anna Rumshisky. 2021.
\newblock \href {https://doi.org/10.18653/v1/2021.findings-acl.300} {{BERT} busters: Outlier dimensions that disrupt transformers}.
\newblock In \emph{Findings of the Association for Computational Linguistics: ACL-IJCNLP 2021}, pages 3392--3405, Online. Association for Computational Linguistics.

\bibitem[{Kudugunta et~al.(2023)Kudugunta, Caswell, Zhang, Garcia, Choquette-Choo, Lee, Xin, Kusupati, Stella, Bapna, and Firat}]{kudugunta2023madlad400multilingualdocumentlevellarge}
Sneha Kudugunta, Isaac Caswell, Biao Zhang, Xavier Garcia, Christopher~A. Choquette-Choo, Katherine Lee, Derrick Xin, Aditya Kusupati, Romi Stella, Ankur Bapna, and Orhan Firat. 2023.
\newblock \href {https://arxiv.org/abs/2309.04662} {Madlad-400: A multilingual and document-level large audited dataset}.
\newblock \emph{Preprint}, arXiv:2309.04662.

\bibitem[{Kumar et~al.(2024)Kumar, Sumers, Yamakoshi, Goldstein, Hasson, Norman, Griffiths, Hawkins, and Nastase}]{Kumar2024-ma}
Sreejan Kumar, Theodore~R Sumers, Takateru Yamakoshi, Ariel Goldstein, Uri Hasson, Kenneth~A Norman, Thomas~L Griffiths, Robert~D Hawkins, and Samuel~A Nastase. 2024.
\newblock Shared functional specialization in transformer-based language models and the human brain.
\newblock \emph{Nat. Commun.}, 15(1):5523.

\bibitem[{Mahowald et~al.(2024)Mahowald, Ivanova, Blank, Kanwisher, Tenenbaum, and Fedorenko}]{mahowald2024dissociating}
Kyle Mahowald, Anna~A Ivanova, Idan~A Blank, Nancy Kanwisher, Joshua~B Tenenbaum, and Evelina Fedorenko. 2024.
\newblock Dissociating language and thought in large language models.
\newblock \emph{Trends in Cognitive Sciences}.

\bibitem[{Meta(2024)}]{grattafiori2024llama}
Llama Team AI~@ Meta. 2024.
\newblock \href {https://arxiv.org/abs/2407.21783} {The llama 3 herd of models}.
\newblock \emph{Preprint}, arXiv:2407.21783.

\bibitem[{Puccetti et~al.(2022)Puccetti, Rogers, Drozd, and Dell{'}Orletta}]{puccetti-etal-2022-outlier}
Giovanni Puccetti, Anna Rogers, Aleksandr Drozd, and Felice Dell{'}Orletta. 2022.
\newblock \href {https://doi.org/10.18653/v1/2022.findings-emnlp.93} {Outlier dimensions that disrupt transformers are driven by frequency}.
\newblock In \emph{Findings of the Association for Computational Linguistics: EMNLP 2022}, pages 1286--1304, Abu Dhabi, United Arab Emirates. Association for Computational Linguistics.

\bibitem[{Qwen et~al.(2024)Qwen, :, Yang, Yang, Zhang, Hui, Zheng, Yu, Li, Liu, Huang, Wei, Lin, Yang, Tu, Zhang, Yang, Yang, Zhou, Lin, Dang, Lu, Bao, Yang, Yu, Li, Xue, Zhang, Zhu, Men, Lin, Li, Xia, Ren, Ren, Fan, Su, Zhang, Wan, Liu, Cui, Zhang, and Qiu}]{qwen2024qwen25}
Qwen, :, An~Yang, Baosong Yang, Beichen Zhang, Binyuan Hui, Bo~Zheng, Bowen Yu, Chengyuan Li, Dayiheng Liu, Fei Huang, Haoran Wei, Huan Lin, Jian Yang, Jianhong Tu, Jianwei Zhang, Jianxin Yang, Jiaxi Yang, Jingren Zhou, Junyang Lin, Kai Dang, Keming Lu, Keqin Bao, Kexin Yang, Le~Yu, Mei Li, Mingfeng Xue, Pei Zhang, Qin Zhu, Rui Men, Runji Lin, Tianhao Li, Tingyu Xia, Xingzhang Ren, Xuancheng Ren, Yang Fan, Yang Su, Yichang Zhang, Yu~Wan, Yuqiong Liu, Zeyu Cui, Zhenru Zhang, and Zihan Qiu. 2024.
\newblock \href {https://arxiv.org/abs/2412.15115} {Qwen2.5 technical report}.
\newblock \emph{Preprint}, arXiv:2412.15115.

\bibitem[{Schilling et~al.(2021)Schilling, Maier, Gerum, Metzner, and Krauss}]{SchillingMGMK21}
Achim Schilling, Andreas Maier, Richard Gerum, Claus Metzner, and Patrick Krauss. 2021.
\newblock \href {https://doi.org/10.1016/j.neunet.2021.03.035} {Quantifying the separability of data classes in neural networks}.
\newblock \emph{Neural Networks}, 139:278--293.

\bibitem[{Stolfo et~al.(2023)Stolfo, Belinkov, and Sachan}]{stolfo-etal-2023-mechanistic}
Alessandro Stolfo, Yonatan Belinkov, and Mrinmaya Sachan. 2023.
\newblock \href {https://doi.org/10.18653/v1/2023.emnlp-main.435} {A mechanistic interpretation of arithmetic reasoning in language models using causal mediation analysis}.
\newblock In \emph{Proceedings of the 2023 Conference on Empirical Methods in Natural Language Processing}, pages 7035--7052, Singapore. Association for Computational Linguistics.

\bibitem[{Tang et~al.(2024)Tang, Luo, Huang, Zhang, Wang, Zhao, Wei, and Wen}]{tang-etal-2024-language}
Tianyi Tang, Wenyang Luo, Haoyang Huang, Dongdong Zhang, Xiaolei Wang, Xin Zhao, Furu Wei, and Ji-Rong Wen. 2024.
\newblock \href {https://doi.org/10.18653/v1/2024.acl-long.309} {Language-specific neurons: The key to multilingual capabilities in large language models}.
\newblock In \emph{Proceedings of the 62nd Annual Meeting of the Association for Computational Linguistics (Volume 1: Long Papers)}, pages 5701--5715, Bangkok, Thailand. Association for Computational Linguistics.

\bibitem[{Team(2024)}]{gemmateam2024gemma2improvingopen}
Gemma Team. 2024.
\newblock \href {https://arxiv.org/abs/2408.00118} {Gemma 2: Improving open language models at a practical size}.
\newblock \emph{Preprint}, arXiv:2408.00118.

\bibitem[{Tenney et~al.(2019)Tenney, Xia, Chen, Wang, Poliak, McCoy, Kim, Durme, Bowman, Das, and Pavlick}]{tenney2018what}
Ian Tenney, Patrick Xia, Berlin Chen, Alex Wang, Adam Poliak, R~Thomas McCoy, Najoung Kim, Benjamin~Van Durme, Sam Bowman, Dipanjan Das, and Ellie Pavlick. 2019.
\newblock \href {https://openreview.net/forum?id=SJzSgnRcKX} {What do you learn from context? probing for sentence structure in contextualized word representations}.
\newblock In \emph{International Conference on Learning Representations}.

\bibitem[{Warstadt and Bowman(2022)}]{warstadt2022what}
Alex Warstadt and Samuel~R Bowman. 2022.
\newblock What artificial neural networks can tell us about human language acquisition.
\newblock In Shalom Lappin and Jean-Philippe Bernardy, editors, \emph{Algebraic {Structures} in {Natural} {Language}}, pages 17--60. CRC Press.

\bibitem[{Weber et~al.(2024)Weber, Jumelet, Bruni, and Hupkes}]{weber-etal-2024-interpretability}
Lucas Weber, Jaap Jumelet, Elia Bruni, and Dieuwke Hupkes. 2024.
\newblock \href {https://doi.org/10.18653/v1/2024.acl-long.248} {Interpretability of language models via task spaces}.
\newblock In \emph{Proceedings of the 62nd Annual Meeting of the Association for Computational Linguistics (Volume 1: Long Papers)}, pages 4522--4538, Bangkok, Thailand. Association for Computational Linguistics.

\bibitem[{Wolf et~al.(2020)Wolf, Debut, Sanh, Chaumond, Delangue, Moi, Cistac, Rault, Louf, Funtowicz, Davison, Shleifer, von Platen, Ma, Jernite, Plu, Xu, Le~Scao, Gugger, Drame, Lhoest, and Rush}]{wolf-etal-2020-transformers}
Thomas Wolf, Lysandre Debut, Victor Sanh, Julien Chaumond, Clement Delangue, Anthony Moi, Pierric Cistac, Tim Rault, Remi Louf, Morgan Funtowicz, Joe Davison, Sam Shleifer, Patrick von Platen, Clara Ma, Yacine Jernite, Julien Plu, Canwen Xu, Teven Le~Scao, Sylvain Gugger, Mariama Drame, Quentin Lhoest, and Alexander Rush. 2020.
\newblock \href {https://doi.org/10.18653/v1/2020.emnlp-demos.6} {Transformers: State-of-the-art natural language processing}.
\newblock In \emph{Proceedings of the 2020 Conference on Empirical Methods in Natural Language Processing: System Demonstrations}, pages 38--45, Online. Association for Computational Linguistics.

\bibitem[{Wu et~al.(2024)Wu, Yu, Yogatama, Lu, and Kim}]{wu2024semantic}
Zhaofeng Wu, Xinyan~Velocity Yu, Dani Yogatama, Jiasen Lu, and Yoon Kim. 2024.
\newblock The semantic hub hypothesis: Language models share semantic representations across languages and modalities.
\newblock \emph{arXiv preprint arXiv:2411.04986}.

\bibitem[{Yu et~al.(2025)Yu, Wang, Reed, and Wan}]{yu2025the}
Mengxia Yu, De~Wang, Colorado Reed, and Alvin Wan. 2025.
\newblock \href {https://openreview.net/forum?id=0Ag8FQ5Rr3} {The super weight in large language models}.

\bibitem[{Zhang et~al.(2023)Zhang, Liu, and Shao}]{zhang-etal-2023-fine}
Zhong Zhang, Bang Liu, and Junming Shao. 2023.
\newblock \href {https://doi.org/10.18653/v1/2023.acl-long.95} {Fine-tuning happens in tiny subspaces: Exploring intrinsic task-specific subspaces of pre-trained language models}.
\newblock In \emph{Proceedings of the 61st Annual Meeting of the Association for Computational Linguistics (Volume 1: Long Papers)}, pages 1701--1713, Toronto, Canada. Association for Computational Linguistics.

\bibitem[{Zhu et~al.(2025)Zhu, Dai, and Sui}]{zhu-etal-2025-language}
Fangwei Zhu, Damai Dai, and Zhifang Sui. 2025.
\newblock \href {https://aclanthology.org/2025.coling-main.47/} {Language models encode the value of numbers linearly}.
\newblock In \emph{Proceedings of the 31st International Conference on Computational Linguistics}, pages 693--709, Abu Dhabi, UAE. Association for Computational Linguistics.

\end{thebibliography}
\end{document}